\documentclass{svproc}
\usepackage{url}

\usepackage{array}
\usepackage{csquotes}

\begin{document}
\mainmatter              
\title{Contextualizing Artificially Intelligent Morality: 
A Meta-Ethnography of Theoretical, Political and Applied Ethics}
\titlerunning{Contextualizing Artificially Intelligent Morality}  
%
\author{Jennafer Shae Roberts\inst{1} \and Laura N Montoya\inst{1}
}
\authorrunning{Roberts and Montoya} 
%
\tocauthor{Jennafer Shae Roberts and Laura N Montoya}
\institute{Accel AI Institute, San Francisco, CA\\
\email{info@accel.ai},\\ 
Home page:
\texttt{http://www.accel.ai}
}

\maketitle              

\begin{abstract}
In this meta-ethnography, we explore three different angles of ethical artificial intelligence (AI) design implementation including the philosophical ethical viewpoint, the technical perspective, and framing through a political lens. Our qualitative research includes a literature review which highlights the cross referencing of these angles through discussing the value and drawbacks of contrastive top-down, bottom-up, and hybrid approaches previously published. The novel contribution to this framework is the political angle, which constitutes ethics in AI either being determined by corporations and governments and imposed through policies or law (coming from the top), or ethics being called for by the people (coming from the bottom), as well as top-down, bottom-up, and hybrid technicalities of how AI is developed within a moral construct and in consideration of its users, with expected and unexpected consequences and long-term impact in the world. There is a focus on reinforcement learning as an example of a bottom-up applied technical approach and AI ethics principles as a practical top-down approach. This investigation includes real-world case studies to impart a global perspective, as well as philosophical debate on the ethics of AI and theoretical future thought experimentation based on historical fact, current world circumstances, and possible ensuing realities. 

\end{abstract}
\section{Introduction}
As a meta-ethnography, this paper will take an anthropological approach to the culture and development of artificial intelligence (AI) ethics and practices. This is in no way exhaustive, however it will magnify some of the key angles and tensions in the ethical AI field. We will be using the previously published \cite{Allen2005ArtificialApproaches} \cite{Wallach2008MachineFaculties} framework of top-down and bottom-up ethics in AI and examining what this means in three different contexts: theoretical, technical, and political. Strategies for artificial morality have been discussed in a top-down, bottom-up and hybrid frameworks in order to create a foundation for philosophical and technical reflections on AI system development. \cite{Allen2005ArtificialApproaches} \cite{Wallach2008MachineFaculties}  Although there can be distinctions made between ethics and morality, we use the terms interchangeably. Top-down ethics in AI can be described as a rule-based system of ethics. These can come from philosophical moral theories (theoretical perspective), from top-down programming (technical perspective), \cite{Allen2005ArtificialApproaches} or by principles designated by authorities (political perspective). Bottom-up ethics in AI is contrastive to Top-down approaches and works without overarching rules. Bottom-up ethics in AI can come from learned experiences (theoretical perspective), from machine learning and reinforcement learning (technical perspective), \cite{Allen2005ArtificialApproaches}  or from everyday people and users of technology calling for ethics (political perspective). Hybrid versions of top-down and bottom-up methods are a mixture of the two, or else somewhere in the middle, and can have various outcomes. The conclusions from this analysis show that ethics around AI is complex, and especially when deployed globally, it's implementation needs to be considered from multiple angles: there is no one correct way to make AI that is ethical. Rather, care must be taken at every turn to right the wrongs of society through the application of AI ethics, which could create a new way of looking at what ethics means in our current digital age.

This paper contextualizes top-down and bottom-up ethics in AI through analyzing theoretical, technical, and political implementations. Section 2 includes a literature review which states our research contributions and how we have built on existing research, as well as an outline of the framework utilized throughout. In section 3, the first angle of the framework described is the theoretical moral ethics viewpoint: Ethics can formulate from the top-down, coming from rules and philosophies, or bottom-up which mirrors the behaviors of people and what is socially acceptable for individuals as well as groups, varying greatly by culture. Section 3.1 gives an example of fairness as a measure of the complexity of theoretical ethics in applied AI. Section 4 regards the technical perspective, which is exemplified by programming from the top and applied machine learning from the bottom: Essentially how to think about implementing algorithmic policies with balanced data that will lead to fair and desirable outcomes. Section 5 will examine the angle of top-down ethics dictated from the powers that be, and bottom-up ethics derived from the demands of the people: We will call this the political perspective. In section 6, we will connect the perspectives all back together and reintegrate these concepts whilst we examine how they intertwine by first looking at the bottom-up method of AI being taught ethics using the example of reinforcement learning in section 6.1. Section 7 combines perspectives from the top-down and illustrates this with section 7.1 that provides examples of principles of AI ethics. An understanding of hybrid ethics in AI which incorporates top-down and bottom-up methods is included in section 8, followed by case studies on data mining in Africa (section 8.1) and the varying efficiency of contact tracing apps for COVID-19 in Korea and Brazil (section 8.2). Following is a discussion in section 9, and finally the paper's conclusion in section 10. This is an exercise in exploration to reach a deeper understanding of morality around AI. How ethics for AI works in reality is a blend of all of these theories and ideas acting on and in conjunction with one another. The aim of this qualitative analysis is to provide a framework to understand and think critically about AI ethics, and our hope is that this will influence the development of ethical AI in the future.

\section{Literature Review}
The purpose of this paper is to deepen the understanding of the development of ethics in AI by cross-referencing a political perspective into the framework of top-down, bottom-up and hybrid approaches, which only previously covered theoretical and technical angles. \cite{Allen2005ArtificialApproaches} Adding the third political perspective fills some of the gaps that were left by only viewing AI ethics theoretically and technically. The political perspective allows for the understanding about where the influence of power affects ethics in AI. Issues of power imbalances are often systemic, and systems of oppression can now be replicated without oversight with the use of AI, as it learns this from past human behaviour, such as favoring white faces and names in algorithms that are used for hiring and loans. \cite{Martin2019EthicalAlgorithms} Furthermore,  it is important to not ignore the fact that AI serves to increase wealth and power for large corporations and governments who have the most political influence. \cite{Whittlestone2019TheEthics} While the influence of politics on the development of AI ethics has been described previously, never has it been discussed in comparison with the technical and theoretical lens while also utilizing the top-down, bottom-up, and hybrid frameworks for development.

By engaging with and expanding on the framework of top-down, bottom-up and hybrid ethics in AI, we can gain a deeper understanding of how ethics could be applied to technology. There has been debate about whether ethical AI is even possible, \cite{Phan2021EconomiesTech} and it is not only due to programming restrictions but to the fact that AI exists in an unethical world which values wealth and power for the people already in power above all else. \cite{Phan2021EconomiesTech} This is why the political perspective is so vital to this discussion. By political, we do not refer to any particular political parties or sides, but rather, as a way of talking about power, both corporate and government. As defined in the Merriam Webster dictionary, political affairs or business refer to ‘competition between competing interest groups or individuals for power and leadership (as in a government). \cite{Merriam-Webster2022Merriam-Webster}

The articles we chose to include in this literature review paint a picture of the trends in AI ethics development in the past two decades. Table 1 features a list of primary research contributions and which frameworks they discuss. The first mention of the framework we use to talk about ethics in AI from the top-down, bottom-up and a mix of the two (hybrid) originated with Allen, Smit and Wallach in 2005. \cite{Allen2005ArtificialApproaches} The same authors penned a second article on this framework for machine morality in 2008. \cite{Wallach2008MachineFaculties} The authors described two different angles of approaching AI morality as being the theoretical angle and the technical angle (but not the political angle). In 2017, \cite{Etzioni2017IncorporatingIntelligence} an article was written that explored humanity’s ethics and the influence on AI ethics, for better or worse. Although the authors did not acknowledge politics or power directly, they did discuss the difference between legality and personal ethics. Next we see a great deal of focus on AI principles, as exemplified by the work of Whittlestone et al in 2019, \cite{Whittlestone2019TheEthics} which also discusses political tensions and other ethical tensions in principles of AI ethics outright. This is important as it questions where the power is situated, as is a central tenet to our addition to this area of research. These papers which describe the political lens and tensions of power and political influence are lacking in the framework of top-down, bottom-up, and hybrid ethical AI implementations.

\begin{table}
  \caption{Literature Contributions including Frameworks Cross-referenced 
}  
  \setlength{\tabcolsep}{8pt}
  \label{tab:princ}
  \begin{tabular}{p{0.2\linewidth} p{0.28\linewidth} p{0.35\linewidth}}
    
    \hline
    Authors and year&Contribution&Frameworks Cross-referenced\\
    \hline
   
     Allen, Smit, \& Wallach (2005) & First paper to use top-down/bottom-up/
hybrid framework for describing artificial morality 

& Theoretical and technical perspectives alongside the top-down, bottom-up, and hybrid framework\\
\hline
    Wallach, Allen, \& Smit (2008) & Values and limitations of top-down imposition and bottom-up building of ethics in AI 

&Theoretical and technical perspectives alongside the top-down, bottom-up approaches
\\
\hline
    Etzioni \& Etzioni (2017)
&Ethical choices of humans over millennia as a factor in building ethical AI&Technical applied development, and political perspectives described but without the top-down, bottom-up framework\\
\hline
    Whittlestone, Nyrup, Alexandrova, \& Cave (2019) & Tensions within principles of AI ethics& Theoretical and political lens for ethical AI development without the technical perspective and without mention of the top-down, bottom-up, and hybrid development framework\\
\hline
    Phan, Goldenfein, Mann, \& Kuch (2021)&Questions the ethics of Big Tech taking responsibility and addresses ethical crises& Political lens for ethical AI development without the technical or theoretical perspectives and without mention of the top-down, bottom-up, and hybrid development framework\\
    
  \hline
 
\end{tabular}
\end{table}

Our methodology utilizes the top-down, bottom-up, and hybrid framework for the development of AI ethics and cross-references it with the technical, theoretical, and political lenses which we will explore more deeply in the following sections. Table 2 lists these cross-references in simplified terms for the reader. 

\begin{table}
  \caption{Framework for Contextualizing AI Ethics 
}
 \setlength{\tabcolsep}{7pt}
  \label{tab:princ}
  \begin{tabular}{p{0.12\linewidth} p{0.22\linewidth} p{0.2\linewidth} p{0.3\linewidth}}
    \hline
    Approach&Top-Down&Bottom-Up&Hybrid\\
    \hline
     Technical & Programmed rules 
Ex: Chatbot & Machine learning 
Ex: Reinforcement learning & Has a base of rules or instructions, but then also is fed data to learn from as it goes
Ex: Autonomous vehicles employ some rule-based ethics, while also learning from other drivers and road experience\\
\hline
    Theoretical&Rule-utilitarianism and deontological ethics, principles, ie fairness
Ex: consequentialist ethics, Kant’s moral imperative and other duty based theories&Experience-based, case-based reasoning
Ex: learn as you go from real-world consequences&Personal moral matrix combining rules and learned experience 
Ex: having some rules but also developing ethics through experiences
\\
\hline
    Political&Corporate and political powers
Ex: corporate control over what is ethical for the company&People power
Ex: groups online of individuals calling for ethics in AI&A combination of ethics from those in power who take into account ethics called for by the people
Ex: employees collaborating with their corporation on ethical AI issues
\\
  \hline
\end{tabular}
\end{table}

\section{Theoretical AI Morality: Top-Down vs Bottom-Up}
The first area to consider is ethics from a theoretical moral perspective. The primary point to mention in this part of the analysis is that ethics has been historically made for people, and people are complex in how they understand and apply ethics, especially top-down ethics. At an introductory glance, “top-down” ethical theories amount to rule-utilitarianism and deontological ethics, where “bottom-up” refers to case-based reasoning. \cite{vanRysewyk2015AData} Theoretical ethics from a top-down perspective includes examples such as the Golden Rule, the Ten Commandments, consequentialist or utilitarian ethics, Kant’s moral imperative and other duty based theories, Aristotle's virtues, and Asimov’s laws of robotics. These are coming from a wide range of perspectives, including literature, philosophy, and religion. \cite{Wallach2008MachineFaculties} Most of these collections of top-down ethical principles are made solely for humans. The one exception on this list that does not apply to people is of course Asimov’s laws of robotics, that are applied precisely for AI. However, Asimov himself said that they were flawed. 

Azimov used storytelling to demonstrate that his three laws, plus the ‘zeroth’ law added in 1985, had problems of prioritization and potential deadlock. He showed that ultimately, the laws would not work, despite the surface-level appearance of putting humanity’s interest above that of the individual. This has been echoed by other theorists on any rule based system implemented for ethical AI. Asimov’s rules of robotics seemingly encapsulate a wide array of ethical concerns, giving the impression of being intuitive and straightforward. However, in each of his stories, we can see how they fail time after time. \cite{Allen2005ArtificialApproaches}  Much of science fiction does not predict the future as much as warn us against its possibilities.

The top-down rule-based approach to ethics presents different challenges for AI than in systems that were originated for humans. As humans, we learn ethics as we go, from observation of our families and community, including how we react to our environment and how others react to us. Etzioni and Etzioni \cite{Etzioni2017IncorporatingIntelligence} made the case that humans first acquire moral values from those who raise them, although it can be argued that individuals make decisions based on their chosen philosophies. As people are exposed to various inputs from new groups, cultures and subcultures, they modify their core value systems, gradually developing their own personal moral matrix. \cite{Etzioni2017IncorporatingIntelligence} This personal moral mix could be thought of as a hybrid model of ethics for humans. The question is, how easy and practical is it to take human ethics and apply them to machines?

Some would say it is impossible to teach AI right and wrong, if we could even come to an agreement on how to define those terms in the first place. Researchers have stressed the importance of differences in the details of ethical systems across cultures and between individuals, even though shared values exist that transcend cultural differences. \cite{Allen2005ArtificialApproaches} There simply is not one set of ethical rules that will be inclusive for everyone. 

\subsection{The Example of Fairness in AI
}

Many of the common systems of values that experts agree need to be considered in AI include fairness, or to take it further, justice. We will work with this concept as an example. We have never had a fair and just world, so to teach an AI to be fair and just does not seem possible. But what if it was? We get into gray areas when we imagine the open-ended potential future of AI. Imagining it could actually improve the state of the world, as opposed to imagining how it could lead to further destruction of humanity could be what propels us in a more positive direction. 

Artificial intelligence should be fair. The first step is to agree on what a word like fairness means when designing AI. Many people pose the question: Fairness for whom? 

Then there is the question of how to teach fairness to AI. AI systems as we know are machines. Machines are good at math. Mathematical fairness and social fairness are two very different things. How can this be codified? Can an equation which solves or tests for fairness between people be developed?

Most AI is built to solve problems of convenience and to automate tedious or monotonous tasks in order to free up our time and make more money. There is a disconnect between what Allen et al. \cite{Allen2005ArtificialApproaches} refer to as the spiritual worldviews from which much of our ethical understanding originates, and the materialistic worldview of computers and robots, which is not completely compatible. 

It can be seen every day in the algorithms that discriminate and codify what retailers think we are most likely to consume. For example, in the ads they show us as we scroll, we often see elements that don’t align with our values but rather appeal to our habits of consumption. At the core level, these values are twisted to benefit the current capitalistic systems and have little to do with actually improving our lives. We cannot expect AI to jump from corporate materialism to social justice, or reach a level of fairness, simply by tweaking the algorithms.

Teaching ethics to AI is extremely challenging -- if not impossible -- on multiple fronts. In order to have ethical AI we need to first evaluate our own ethics. Douglas Rushkoff, well-known media theorist, author and Professor of Media at City University of New York, wrote: 

\begin{displayquote}
“...the reason why I think AI won’t be developed ethically is because AI is being developed by companies looking to make money – not to improve the human condition. . . My concern is that even the ethical people still think in terms of using technology on human beings instead of the other way around. So, we may develop a ‘humane’ AI, but what does that mean? It extracts value from us in the most ‘humane’ way possible?” \cite{Rainie2021ExpertsDecade.} 
\end{displayquote}

This is a major consideration for AI ethics, and the realities of capitalism don't align with ethics and virtues such as fairness. One of the biggest questions when considering ethics for AI is how to implement something so complex and philosophical into machines that are contrastingly good at precision. Some say this is impossible. “Ethics is not a technical enterprise, there are no calculations or rules of thumb that we could rely on to be ethical. Strictly speaking, an ethical algorithm is a contradiction in terms.” (Vachnadze, 2021) \cite{Vachnadze2021ReinforcementMachines.} The potential possibilities for technical application of AI morality.

\section{Technical AI Morality: Top-Down vs Bottom-Up}

One way to think about top-down AI from the technical perspective, as noted by Eckart, \cite{Eckart2020Top-downAI} is to think of it as a decision tree, often implemented in the form of a call center chat bot. The chat bot guides the user through a defined set of options depending on the answers inputted. Eckart continues by describing bottom-up AI as what we typically think of when we hear artificial intelligence: utilizing machine learning and deep learning. As an example, we can think about the AI utilized for diagnostic systems in healthcare and self-driving cars. These bottom-up systems can learn automatically without explicit programming from the start. \cite{Eckart2020Top-downAI}

Top-down systems of learning can be very useful for some tasks that machines can be programmed to do, like the chatbot example above. However, if they are not monitored, they could make mistakes, and it is up to us as people to catch those mistakes and correct them, which is not always possible with black boxes in effect. There may also be a lack of exposure to sufficient data to make a decision or prediction in order to solve a problem, leading to system failure. Here is the value of having a ‘human in the loop’. This gets complicated even further when we move into attempting to program the more theoretical concepts of ethics.

Bottom-up from the technical perspective, which will be described in depth below, follows the definition of machine learning. The system is given data to learn from, and it uses that information from the past to predict and make decisions for the future. This can work quite well for many tasks. This approach can also have a lot of flaws built in, because the world that it learns from is flawed. We can look at the classic example of harmful biases being learned and propagated through a system, for instance in who gets a job or a loan, because the data from the past reflects biased systems in our society. \cite{Martin2019EthicalAlgorithms}

Technical top-down and bottom-up ethics in AI primarily concerns how AI learns ethics. Machines don’t learn like people do. They learn from the data that is fed to them, and they are very good at certain narrow tasks, such as memorization or data collection. However, AI systems can fall short in areas such as objective reasoning, which is at the core of ethics. Whether coming from the top-down or bottom-up, the underlying concern is that teaching ethics to AI is extremely difficult, both technically and socially. 

Ethics and morality prove difficult to come to an actual consensus on. We live in a very polarized world. What is fair to some will undoubtedly be unfair to others. There are several hurdles to overcome. Wallach et al \cite{Wallach2008MachineFaculties} describe three specific challenges to address in this matter, paraphrased below:

\begin{enumerate}
    \item Scientists must break down moral decision making into its component parts, which presents an engineering task of building autonomous systems in order to safeguard basic human values.
    \item Which decisions can and cannot be codified and managed by mechanical systems needs to be recognized.
    \item Designing effective and cognitive systems which are capable of managing ambiguity and conflicting perspectives needs to be learned. \cite{Wallach2008MachineFaculties}
\end{enumerate}

Here we will include the use of a hybrid model of top-down and bottom-up ethics for AI, that has a base of rules or instructions, but then also is fed data to learn from as it goes. This method claims to be the best of both worlds, and covers some of the shortcomings of both top-down and bottom-up models. For instance, self-driving cars can be programmed with laws and rules of the road, and also can learn from observing human drivers. In the next section we will explore more of the political angle of this debate. 

\section{Political AI Morality: Top-Down vs Bottom-Up
}

We use the term political to talk about where the power and decision making is coming from, which then has an effect that radiates outward and influences systems, programmers, and users alike. As an example of top-down from a political perspective, this paper will largely concern itself with principles of ethics in AI, often stated by corporations and organizations. Bottom-up ethics in AI from a political standpoint concerns the perspectives of individuals and groups who are not in positions of power, yet still need a voice. 

The Asilomar AI Principles \cite{2022AIInstitute}  are an example of a top-down model and have their critiques. This is a comprehensive list of rules that was put out by the powers that be in tech and AI, with the hopes of offering guidelines for developing ethics in AI. Published in 2017, this is one key example of top-down ethics from the officials including 1,797 AI/Robotics Researchers and 3,923 other Endorsers affiliated with the Future of Life Institute. These principles outline ethics and values that the use of AI must respect, provide guidelines on how research should be conducted, and offer important considerations for thinking about long-term issues. \cite{2022AIInstitute} Congruently, another set of seven principles for Algorithmic Transparency and Accountability were published by the US Association for Computing Machinery (ACM) which addressed a narrower but closely related set of issues. \cite{Whittlestone2019TheEthics}  Since then we have seen an explosion of lists of principles for AI ethics. A deeper discussion of principles can be found in section 7.1 of this paper. 

The bottom-up side of the political perspective is not as prevalent but could look like crowd-collected considerations about ethics in AI, such as from employees at a company, students on a campus, or online communities. The key feature of bottom-up ethics from a political perspective is determinism by everyday people, mainly, the users of the technology. MIT’s moral machine (which collected data from millions of people on their decisions in a game-like program to assess what a self-driving vehicle should do in life or death situations), is one example of this. \cite{Awad2018TheExperiment} However, it still has top-down implications such as obeying traffic laws imposed by municipalities. A pure bottom-up community-driven ethics initiative could include guidelines, checklists, and case studies specific to the ethical challenges of crowdsourced tasks. \cite{Shmueli2021BeyondCrowdsourcing} 

Even when utilizing bottom-up “crowdsourcing” and employing the moral determination of the majority, these systems often fail to serve minority participants. In a roundtable discussion from the Open Data Initiative (ODI), \cite{ExperimentalismDocs.}  they found that marginalized communities have a unique placement for understanding and identifying the contradictions and tensions of the systems we all operate in. Their unique perspectives could be leveraged to create change. If a system works for the majority, which is often the goal, it may be invisibly dysfunctional for people outside of the majority. This insight is invaluable to alleviate ingrained biases. 

There is an assumption that bottom-up data institutions will represent everyone in society and always be benign. Alternatively, there is a counter-argument that their narrow focus leads to niche datasets and lacks applicability to societal values. In the best light, bottom-up data institutions are viewed as revolutionary mechanisms that could rebalance power between big tech companies and communities. \cite{ExperimentalismDocs.}  An important point to keep in mind when thinking about bottom-up ethics is that there will always be different ideals coming from different groups of people, and the details of the applications are where the disagreements abound. 

\section{The Bottom-up Method of AI Being Taught Ethics through Reinforcement Learning }

To recombine the perspectives of theoretical, technical, and political bottom-up ethics for AI is a useful analytical thought experiment. Allen et. al. \cite{Allen2005ArtificialApproaches} describe bottom-up approaches to ethics in AI as those which learn through experience and strive to create environments where appropriate behavior is selected or rewarded, instead of functioning under a specific moral theory. These approaches learn either by unconscious mechanistic trial and failure of evolution, by engineers or programmers adjusting to new challenges they encounter, or by the learning machine’s own educational development. \cite{Allen2005ArtificialApproaches} The authors explain the difficulties of evolving and developing strategies that hold the promise of a rise in skills and standards that are integral to the overall design of the system. Trial and error are the fundamental tenets of evolution and learning, which rely heavily on learning from unsuccessful strategies and mistakes. Even in the fast-paced world of computer processing and evolutionary algorithms, this is an extremely time consuming process. Additionally, we need safe spaces for these mistakes to be made and learned from, where ethics can be developed without real-world consequences. 

\subsection{Reinforcement Learning as a Methodology for Teaching AI Ethics
}

Reinforcement learning (RL) is a technique of machine learning where an agent learns by trial and error in an interactive environment, utilizing feedback from its own actions and experiences. \cite{Bhatt2018Reinforcement101} Reinforcement learning is different from other forms of learning that rely on top-down rules. Rather, this system learns as it goes, making many mistakes but learning from them, and it adapts through sensing the environment. RL is commonly used in training algorithms to play games, such as Alpha Go and chess. When it originated, RL was studied in animals, as well as early computers. The trial and error beginnings of this technique have origins in the psychology of animal learning in the early 1900s (Pavlov), as well as in some of the earliest work in AI. This coalesced in the 1980s to develop into the modern field of reinforcement learning. \cite{Sutton2018ReinforcementEdition}

RL utilizes a goal-oriented approach, as opposed to having explicit rules of operation. A ‘rule’ in RL can come about as a temporary side-effect as it attempts to solve the problem, however if the rule proves ineffective later on, it can be discarded. The function of RL is to compensate for machine learning draw-backs by mimicking a living organism as much as possible. \cite{Vachnadze2021ReinforcementMachines.}  This style of learning that throws the rule book out the window could be promising for something like ethics, where the rules are not overly consistent or even agreed upon. Ethics is more situation-dependent, therefore teaching a broad rule is not always sufficient. It is a worthwhile investigation to question if RL could be methodized in integrating ethics into AI. 

The problems addressed by RL consist of learning what to do and how to navigate situations into actions in order to maximize a numerical reward signal. The three most important distinguishing features of RL are: First, that it is essentially a closed-loop; second, that it is not given direct instructions on what actions to take; and third, that there are consequences (reward signals) playing out over extended periods of time \cite{Sutton2018ReinforcementEdition}.
Turning ethics into numerical rewards can pose many challenges, but may be a hopeful consideration for programming ethics into AI systems. Critically, the agent must be able to sense its environment to some degree and it must be able to take actions that affect the state. \cite{Sutton2018ReinforcementEdition}

One of the ways that RL can work in an ethical sense, and to avoid pitfalls, is by utilizing systems that keep a human in the loop. “Interactive learning constitutes a complementary approach that aims at overcoming these limitations by involving a human teacher in the learning process.” \cite{Najar2021ReinforcementSurvey} Keeping a human in the loop is critical for many issues, including those around transparency. Moral uncertainty needs to be considered, purely because ethics is an area of vast uncertainty, and is not an answerable math problem with predictable results. \cite{Ecoffet2020ReinforcementUncertainty} Could an RL program eventually learn how to compute all the different ethical possibilities? 

This may take a lot of experimentation. It is important to know the limitations, while also remaining open to being surprised. We worry a lot about the unknowns of AI: Will it truly align with our values? Only through experimentation can we find out. Researchers stress the importance of RL systems needing a ‘safe learning environment’ where they can learn without any harm being caused to humans, assets, or the external environment. The gap between simulated and actual environments, however, complicates this issue, particularly related to differentiating societal and human values. \cite{Bragg2018WhatLearning}

\section{The Top-Down Method of AI Being Taught Ethics}

Summarizing top-down ethics for AI brings together the philosophical principles, programming rules, and authoritative control in this area. A common thread among all sets of top-down principles is ensuring AI is used for “social good” or “the benefit of humanity”. These phrases carry with them few if any real commitments, hence, a great majority of people can agree on them. However, many of these proposed principles for AI ethics are simply too broad to be action guiding. \cite{Whittlestone2019TheEthics}  Furthermore, if these principles are being administered from big Tech or the government in a top-down political manner, there could be a lot that slips under the radar because it sounds good. Relating to the earlier example in section 2.1, ‘fairness’ is something we can all agree is good, but we can’t all agree what it means. Fair for one person or group could equate to really unfair to another. 

According to Wallach et. al, \cite{Wallach2008MachineFaculties} the price of top-down theories can amount to static definitions which fail to accommodate new conditions, or may potentially be hostile. The authors note that the meaning and application of principle goals can be subject to debate due to them being overly vague and abstract. \cite{Wallach2008MachineFaculties} This is a problem that will need to be addressed going forward. A machine doesn’t implicitly know what ‘fairness’ means. So how can we teach it a singular definition when fairness holds a different context for everyone? Next, we turn to the area of principles of AI ethics to explore the top-down method further.

\subsection{Practical Principles for AI Ethics 
}

Principles of AI are a top-down approach to ethics for artificial intelligence. In the last few years, we have been seeing lists of principles for AI ethics emerging prolifically. These lists are very useful, not only for AI and its impact, but also on a larger social level. Because of AI, people are thinking about ethics in a whole new way: How do we define and digest ethics in order to codify it? 

Principles can be broken into two categories: principles for people who program AI systems to follow, and principles for the AI itself. Some of the principles for people, mainly programmers and data scientists, read like commandments. For instance, The Institute for Ethical AI and ML \cite{TheInstituteforEthicalAiMachineLearning2021TheSystems.} has a list of eight principles geared toward technologists that can be viewed in Table 3.

\begin{table}
  \caption{Principles and their commitments for technologists to develop machine learning systems responsibly as described in the practical framework to develop AI responsibly by The Institute for Ethical AI {\&} Machine Learning \cite{TheInstituteforEthicalAiMachineLearning2021TheSystems.}}
  \label{tab:princ}
  \begin{tabular}{p{0.30\linewidth} p{0.6\linewidth}}
    \hline
    Principle&Commitment of Technologists\\
    \hline
    Human Augmentation & to keep a human in the loop\\ \hline
    Bias Evaluation & to continually monitor bias\\\hline
    Explainability and Justification & to improve transparency\\\hline
    Reproducibility & to ensure infrastructure that is reasonably reproducible\\\hline
    Displacement strategy & to mitigate impact on workers due to automation\\\hline
    Practical accuracy & to align with domain-specific applications\\\hline
    Trust by privacy & to protect and handle data\\\hline
    Data risk awareness & to consider data and model security \\ 
  \hline
\end{tabular}
\end{table}

Other lists of principles are geared towards the ethics of AI systems themselves and what they should adhere to. One such list consists of four principles, published by the National Institute of Standards and Technology (NIST) \cite{Phillips2021FourIntelligence} and are intended to promote explainability. These can be viewed in Table 4.

\begin{table}
  \caption{{Principles and their commitments for responsible 
machine learning and AI systems} \cite{Phillips2021FourIntelligence}}
  \label{tab:princ}
  \begin{tabular}{p{0.30\linewidth} p{0.6\linewidth}}
    \hline
   Principle& Commitment of AI System\\
    \hline
    Explanation  &     provide evidence and reasons for its processes and outputs\\\hline
     Meaningful and understandable  &    have methods to evaluate meaningfulness\\\hline
    Explanation accuracy  &   correctly reflect the reason(s) its generated output\\\hline
    Knowledge limits .  & ensure that a system only operates under conditions for which it was designed and that it does not give overly confident answers in areas it has limited knowledge of. ex: a system programmed to classify birds being used to classify an apple.\\

  \hline
\end{tabular}
\end{table}

Many of the principles overlap across corporations and agencies. A detailed graphic and writeup published by the Berkman Klein Center for Internet and Society at Harvard gives a detailed overview of  forty seven principles that various organizations, corporations, and other entities are adopting, including where they overlap and their definitions. The authors provide many lists and descriptions of ethical principles for AI, and categorize them into eight thematic trends, listed on the following page: 

\begin{enumerate}
    \item Privacy
    \item Accountability
    \item Safety and security
    \item Transparency and explainability
    \item Fairness and non-discrimination
    \item Human control of technology
    \item Professional responsibility
    \item Promotion of human values \cite{Fjeld2020PrincipledAI}
    \end{enumerate}

One particular principle that is missing from these lists regards taking care of the natural world and non-human life. As Boddington states in her book, Toward a Code of Ethics for Artificial Intelligence (2018), “. . . we are changing the world, AI will hasten these changes, and hence, we’d better have an idea of what changes count as good and what count as bad.” \cite{Boddington2017TowardsIntelligence} We will all have different opinions on this, but it needs to be part of the discussion. We can’t continue to destroy the planet while trying to create super AI, and still be under the illusion that our ethics principles are saving the world. 
 
Many of these principles are theoretically sound, yet act as a veil that presents the illusion of ethics. This can be dangerous because it makes us feel like we are practicing ethics while business carries on as usual. Part of the reason for this is because the field of ethical AI development is so new and more research must be done yet to ensure the overall impact is a benefit to society. “Despite the proliferation of these ‘AI principles,’ there has been little scholarly focus on understanding these efforts either individually or as contextualized within an expanding universe of principles with discernible trends.” \cite{Fjeld2020PrincipledAI}

Principles are a double sided coin. On one hand, making the stated effort to follow a set of ethical principles is good. It is beneficial for people to be thinking about doing what is right and ethical, and not just blindly entering code that could be detrimental in unforeseen ways. Some principles are simple in appearance yet incredibly challenging in practice. For example, if we look at the commonly adopted principle of transparency, there is quite a difference between saying that algorithms and machine learning should be explainable and actually developing ways to see inside of the black box. As datasets get bigger, this presents more and more technical challenges. \cite{Boddington2017TowardsIntelligence} Furthermore, some of the principles can conflict with each other, which can land us in a less ethical place than where we started. For example, transparency can conflict with privacy, another popular principle. We can run into a lot of complex problems around this, which needs to be addressed quickly and thoroughly as we move into the future.
 
Overall, we want these concepts in people's minds: such as Fairness. Accountability, and Transparency. These are the core tenets and namesake of the FAccT conference \cite{2022ACMConference} that addresses these principles in depth. It is incredibly important for corporations and programmers to be concerned about the commonly addressed themes of bias, discrimination, oppression, and systemic violence. Yet, what can happen is that these principles make us feel like we are doing the right thing, however, how much does writing out these ideals actually change things? 

In order for AI to be ethical, A LOT has to change, and not just in the tech world. There seems to be an omission of the unspoken principles: the value of money for corporations and those in power and convenience for those who can afford it. If we are aiming to create fairness, accountability, and transparency in AI, we need to do some serious work on society to adjust our core principles away from money and convenience and towards taking care of everyone’s basic needs and the Earth. 

Could AI be a tool that has a side effect of starting an ethics revolution? 

How do we accomplish this? The language that we use is important, especially when it comes to principles. Moss and Metcalf pointed out the importance of using market-friendly terms. If we want morality to win out, we need to justify the organizational resources necessary, when more times than not, companies will choose profit over social good. \cite{Moss2019TheCompanies.} Whittlestone et al. describe the need to focus on areas of tension in ethics in AI, and point out the ambiguity of terms like  ‘fairness’, ‘justice’, and ‘autonomy’. The authors prompt us to question how these terms might be interpreted differently across various groups and contexts. \cite{Whittlestone2019TheEthics} They go on to say that principles need to be formalized into standards, codes and ultimately regulation in order to be useful in practice. Attention is drawn to the importance of acknowledging tensions between high-level goals of ethics, which can differ and even contradict each other. In order to be effective,  it is vital to include a measure of guidance on how to resolve different scenarios. In order to reflect genuine agreement, there must be acknowledgement and accommodation of different perspectives and values as much as possible. \cite{Whittlestone2019TheEthics} The authors then introduce four reasons that discussing tensions is beneficial and important for AI ethics: 

\begin{enumerate}
    \item Bridging the gap between principles and practice
    \item Acknowledging differences in values
    \item Highlighting areas where new solutions are needed 
    \item  Identifying ambiguities and knowledge gaps \cite{Whittlestone2019TheEthics}
   
    \end{enumerate}

Each of these needs to be considered ongoing, as these tensions don’t get solved overnight. Particularly, creating a bridge between principles and practice is important. 
\begin{displayquote}
“We need to balance the demand to make our moral reasoning as robust as possible, with safeguarding against making it too rigid and throwing the moral baby out with the bathwater by rejecting anything we can’t immediately explain. This point is highly relevant both to drawing up codes of ethics, and to the attempts to implement ethical reasoning in machines.” \cite{Boddington2017TowardsIntelligence}
\end{displayquote}

Codes of ethics and ethical principles for AI are important and help start important conversations. However, it can’t stop there. The future will see more and more ways that these principles are put into action, and bring technologists and theorists together to investigate ways to make them function efficiently and ethically. We must open minds to ideas beyond making money for corporations and creating conveniences, and rather toward addressing tensions and truly creating a world that works for everyone. 

\section{The Hybrid of Bottom-Up and Top-Down Ethics for AI}

We have reviewed the benefits and flaws of a top-down approach to ethics in AI, and visited the upsides and pitfalls of the bottom-up approach as well. Many argue that the solution lies somewhere in between, in a hybrid model. 

\begin{displayquote}
“If no single approach meets the criteria for designating an artiﬁcial entity as a moral agent, then some hybrid will be necessary. Hybrid approaches pose the additional problems of meshing both diverse philosophies and dissimilar architectures.” \cite{Allen2005ArtificialApproaches}
\end{displayquote}

Many agree that a hybrid of top-down and bottom-up would be the most effective model for ethical AI. Further, some argue that we need to question the ethics of people, both as the producers and consumers of technology, before we can start to assess fairness in AI.

Researchers state that hybrid AI combines the most desirable aspects of bottom-up, such as neural networks, and top-down, also referred to as symbiotic AI. \cite{SAGAR2021WhatAI} When huge data sets are combined, neural networks are allowed to extract patterns. Then, information can be manipulated and retrieved by rule-based systems utilizing algorithms to manipulate symbols. \cite{SAGAR2021WhatAI} Further research has observed the complementary strengths and weaknesses of bottom-up and top-down strategies. Raza et al. developed a hybrid program synthesis approach, improving top-down interference by utilizing bottom-up analysis. \cite{Raza2020WebInference} When we apply this to ethics and values, ethical concerns that arise from outside of the entity are emphasized by top-down approaches, while the cultivation of implicit values arising from within the entity are addressed by bottom-up approaches. \cite{Wallach2008MachineFaculties} While the authors stated that hybrid systems lacking effective or advanced cognitive faculties will be functional across many domains, they noted how essential it is to recognize times when additional capabilities will be required. \cite{Wallach2008MachineFaculties}

Theoretically, hybrid ethic for AI which features the best of top-down and bottom-up methods in combination is incredibly promising, but in reality, many of the semi-functional or non-functional applications of supposed ethical AI prove challenging and have unforeseen side effects. Many real-world examples could be seen as a hybrid of ethics in AI, and not all have beaming qualities of top-down and bottom-up ethics; rather, they represent the messiness and variance of life. Next we will explore a selection of case studies, which will reflect some ethical AI concerns in real-world examples from across the globe. 

\subsection{Data Mining Case Study: The African Indigenous Context
}

Data sharing, or data mining, is a prime example of conflicting principles of AI ethics. On one hand, it is the epitome of transparency and a crucial element to scientific and economic growth. On the other hand, it brings up serious concerns about privacy, intellectual property rights, organizational and structural challenges, cultural and social contexts, unjust historical pasts, and potential harms to marginalized communities. \cite{Abebe2021NarrativesAfrica} We can reflect on this as a hybrid of top-down and bottom-up ethics in AI, since it utilizes top-down politics, bottom-up data collection, and is theoretically a conflict between the principles of the researchers and the researched communities. 

The term data colonialism can be used to describe some of the challenges of data sharing, or data mining, which reflect the historical and present-day colonial practices such as in African and Indigenous contexts. When we use terms such as ‘mining’ to discuss how data is collected from people, the question remains, who benefits from the data collection? The use of data can paradoxically be harmful to communities it is collected from. Trust is challenging due to the historical actions taken by data collectors while mining data from Indigenous populations. What barriers exist that prevent collected data from being of benefit to African people? We must address the entrenched legacies of power disparities concerning what challenges they present for modern data sharing. \cite{Abebe2021NarrativesAfrica} 

One problematic example is of a non-government organization (NGO) that tried to ‘fix’ problems for marginalized ethnic groups and ended up causing more harm than good. In this case, a European-based NGO planned to address the problem of access to clean potable water in Buranda, while simultaneously testing new water accessibility technology and online monitoring of resources. \cite{Abebe2021NarrativesAfrica} The NGO failed to understand the perspective of the community on the true central issues and potential harms. Sharing the data publicly, including geographic locations, put the community at risk, as collective privacy was violated. In the West privacy is often thought of as a personal concern, however collective identity serves as a great importance to a multitude of African and Indigenous communities. This introduced trust issues due to the disempowerment of local communities in the decision-making process. 

Another case study in Zambia observed that up to 90\% of health research funding comes from external funders, meaning the bargaining power gives little room for negotiations for Zambian scholars. In the study, power imbalances were reported in everything from funding to agenda setting, data collection, analysis, interpretation, and reporting of results. \cite{Abebe2021NarrativesAfrica} This example exhibits further the understanding that trust cannot be formed on the foundation of these imbalances of power. 

Due to this lack of trust, many researchers have run into hurdles with collecting data from marginalized communities. Many of these research projects lead with good intentions, yet there was a lack of forethought into the ethical use of data, during and after the project, which can create unforeseen and irreparable harms to the wellbeing of communities. This creates a hostile environment to build relationships of respect and trust. \cite{Abebe2021NarrativesAfrica}

To conclude this case study in data mining, we can pose the ethical question, “is data sharing good/beneficial?” First and foremost, local communities must be the primary beneficiaries of responsible data sharing practices. \cite{Abebe2021NarrativesAfrica}  It is important to specify who benefits from data sharing, and to make sure that it is not doing any harm to the people behind the data.

\subsection{Contact Tracing for COVID-19 Case Study}

Another complex example of ethics in AI can be seen in the use of contact tracing during the COVID-19 pandemic. Contact tracing can be centralized or non-centralized, which directly relates to top-down and bottom-up methods. The centralized approach is what was deplored in South Korea, where by law, and for the purposes of infectious disease control, the national authority is permitted to collect and use information on all COVID-19 patients and their contacts. \cite{Fendos2020PARTReplicate} In 2020, Germany and Israel tried and failed at adopting centralized approaches, due to a lack of exceptions for public health emergencies in their privacy laws. \cite{Fendos2020PARTReplicate} Getting past the legal barriers can be a lengthy and complex process and not conducive to applying a centralized contract tracing system for the outbreak. \cite{Fendos2020PARTReplicate}

Non-centralized approaches to contact tracing are essentially smartphone apps which track proximal coincidence with less invasive data collection methods. These approaches have thus been adopted by many countries, and don’t have the same cultural and political obstacles as centralized approaches, avoiding legal pitfalls and legislative reform. \cite{Fendos2020PARTReplicate}

Justin Fendos, a professor of cell biology at Dongseo University in Busan, South Korea, wrote that in supporting the public health response to COVID-19, Korea had the political willingness to use technological tools to their full potential. \cite{Fendos2020HowResponse} The Korean government had collected massive amounts of transaction data to investigate tax fraud even before the COVID-19 outbreak. Korea’s government databases hold records of literally every credit card and bank transaction, and this information was repurposed during the outbreak to retroactively track individuals. In Korea, 95\% of adults own a smartphone and many use cashless tools everywhere they go, including on buses and subways. \cite{Fendos2020HowResponse} Hence, contact tracing in Korea was extremely effective.

Public opinion about surveillance in Korea has been stated to be overwhelmingly positive. Fatalities in Korea due to COVID-19 were a third of the global average as of April 2020, when it was also said that they were one of the few countries to have successfully flattened the curve. There have been concerns, despite the success, regarding the level of personal details released by health authorities, which have motivated updated surveillance guidelines for sensitive information. \cite{Fendos2020HowResponse}

Turning to the other side of the planet, a very different picture can be painted. One study focused on three heavily impacted cities in Brazil which had the most deaths from COVID-19 until the first half of 2021. The researchers provided a methodology for applying data mining as a public health management tool, including identifying variables of climate and air quality in relation to the number of COVID-19 cases and deaths. They used rules-based forecasting models and provided forecasting models of new COVID-19 cases and daily deaths in the three Brazilian cities studied. (São Paulo, Rio de Janeiro and Manaus) \cite{Barcellos2021DataMetropolis}

However, the researchers noted that counting of cases in Brazil was affected by high underreporting due to low testing, as well as technical and political problems, hence the study stated that cases may have been up to 12 times greater than investigations indicated. \cite{Barcellos2021DataMetropolis} This shows us that the same technology cannot necessarily be scaled to work for all people in all places across the globe, and that individual concern must be taken when looking for the best solutions for everyone. 

\section{Discussion
}
In the primary paper that this research builds on titled Artificial Morality: Top-down, Bottom-up, and Hybrid Approaches, the authors lead by stating: “Artiﬁcial morality shifts some of the burden for ethical behavior away from designers and users, and onto the computer systems themselves,” \cite{Allen2005ArtificialApproaches} This is a questionable claim. Machines cannot be held responsible for what they learn from people, ever. Machines do not have an inherent conscience or morality as humans do. Moreover, AI can act as a mirror, and the problems that arise in AI often reflect the problems we have in society. People need to assume responsibility, both as individuals and as a society at large. Corporations and governments need to cooperate, and individual programmers and technologists should continually question and evaluate these systems and their morality. In this way, we can use AI technology in an effort to improve society, and create a more sustainable world for everyone. 

The approach of moral uncertainty is intriguing because there isn't ever one answer or solution to an ethical question, and to admit uncertainty leaves it open to continued questioning  that can lead us to the answers that may be complex and decentralized. This path could possibly create a system that can adapt to meet the ethical considerations of everyone involved. \cite{Wallach2008MachineFaculties} Ultimately, societal ethics need to be considered, as AI does not exist in a vacuum. A large consideration is technology in service of making money, primarily for big corporations, and not for improving lives and the world. As long as this is the backbone driving AI and other new technology, we cannot reach true ethics in this field. Given our tendency for individualism over collectivism, who gets to decide what codes of ethics AI follows? If it is influenced by Big Tech, which is often the case, it will serve to support the ethics of a company, which generally has the primary goal of making money for that company. The value of profit over all else needs to shift. 

\begin{displayquote}
“Big Tech has transformed ethics into a form of capital — a transactional object external to the organization, one of the many ‘things’ contemporary capitalists must tame and procure. . . By engaging in an economy of virtue, it was not the corporation that became more ethical, but rather ethics that became corporatised. That is, ethics was reduced to a form of capital — another industrial input to maintain a system of production, which tolerated change insofar as it aligned with existing structures of profit-making.” \cite{Phan2021EconomiesTech}
\end{displayquote}

This reflects the case study of data mining in African communities, whose researchers set out to do good, however were still working in old frameworks around mining resources for personal gain, regurgitating colonialism. Until we can break free from these harmful systems, building an ethical AI is either going to continue to get co-opted and re-capitalized, or possibly, it will find a way to create brand new systems where it can truly be ethical, creating a world where other worlds are possible.

 To leave us with a final thought:  “Ethical issues are never solved, they are navigated and negotiated as part of the work of ethics owners.” \cite{Moss2019TheCompanies.}

\section{Conclusion
}
We have explored ethics in AI implementation in three ways:  theoretically, technically, and politically as described through top-down, bottom-up, and hybrid frameworks. Within this paper, we reviewed reinforcement learning as a bottom-up example, and principles of AI ethics as a top-down example. The concept of fairness as a key ethical value for AI was discussed throughout. Case studies were reviewed to exemplify just how complex and variant ethics in AI can be in different cultures and at different times. The conclusion is that ethics in AI needs a lot more research and work, and needs to be considered from multiple angles while being continuously monitored for unforeseen side effects and consequences. Furthermore, societal ethics need to be accounted for. Our hope is that at least for those who are intending to build and deploy ethical AI systems, they will consider all angles and blind spots, including who might be marginalized or harmed by the technology, especially when it aims to help. By continuing to work on the seemingly impossible task of creating ethical AI, this will radiate out to society and ethics will become more and more of a power in itself that can have wider implications for the betterment of all.

%
%
\bibliographystyle{plain}
\bibliography{references.bib}

\end{document}